\def\year{2022}\relax
\documentclass[letterpaper]{article} 
\usepackage{aaai22}  
\usepackage{times}  
\usepackage{helvet}  
\usepackage{courier}  
\usepackage[hyphens]{url}  
\usepackage{graphicx} 
\usepackage{amsmath}
\usepackage{natbib}  
\usepackage{caption} 
\DeclareCaptionStyle{ruled}{labelfont=normalfont,labelsep=colon,strut=off} 
\usepackage{xcolor}
\usepackage[colorlinks=true,citecolor=blue, linkcolor=blue]{hyperref}
\usepackage{algorithm}
\usepackage{algorithmic}

\usepackage{newfloat}
\usepackage{listings}
\floatstyle{ruled}
\newfloat{listing}{tb}{lst}{}
\floatname{listing}{Listing}

\title{Simultaneously Predicting Multiple Plant Traits \\
from Multiple Sensors via Deformable CNN Regression}
\author{
    Pranav Raja \textsuperscript{\rm 1}\textsuperscript{,}\textsuperscript{\rm 3},
    Alex Olenskyj \textsuperscript{\rm 1}\textsuperscript{,}\textsuperscript{\rm 3},
    Hamid Kamangir \textsuperscript{\rm 1}\textsuperscript{,}\textsuperscript{\rm 3},
    Mason Earles \textsuperscript{\rm 1}\textsuperscript{,}\textsuperscript{\rm 2}\textsuperscript{,}\textsuperscript{\rm 3}
}
\affiliations{
    \textsuperscript{\rm 1}Biological and Agricultural Engineering, University of California, Davis\\
    \textsuperscript{\rm 2}Viticulture and Enology, University of California, Davis\\
    \textsuperscript{\rm 3}AI Institute for Food Systems (AIFS)\\
    {\{pvraja, agolenskyj, hkamangir, jmearles\}@ucdavis.edu}

}

\begin{document}

\maketitle

\begin{abstract}
    Trait measurement is critical for the plant breeding and agricultural production pipeline. Typically, a suite of plant traits is measured using laborious manual measurements and then used to train and/or validate higher throughput trait estimation techniques. Here, we introduce a relatively simple convolutional neural network (CNN) model that accepts multiple sensor inputs and predicts multiple continuous trait outputs – i.e. a multi-input, multi-output CNN (MIMO-CNN). Further, we introduce deformable convolutional layers into this network architecture (MIMO-DCNN) to enable the model to adaptively adjust its receptive field, model complex variable geometric transformations in the data, and fine-tune the continuous trait outputs. We examine how the MIMO-CNN and MIMO-DCNN models perform on a multi-input (i.e. RGB + depth images), multi-trait output lettuce dataset from the 2021 Autonomous Greenhouse Challenge. Ablation studies were conducted to examine the effect of using single versus multiple inputs, and single versus multiple outputs. The MIMO-DCNN model resulted in a normalized mean squared error (NMSE) of 0.068; a substantial improvement over the top 2021 leaderboard score of 0.081. Open-source code is provided.
\end{abstract}

\section{Introduction and Related Work}

Multi-input datasets and multi-input modeling problems are common in breeding and agriculture machine learning applications. Examples of such public datasets include RGB + depth \cite{GENEMOLA2019104289, kusumam2016can, akbar2016novel}, RGB + infrared \cite{haug2014crop, sa2017weednet, lottes2018fully}, and RGB + thermal \cite{elsherbiny2021integration}. Such inputs are typically used for prediction tasks such as classification \cite{lashgari2020fusion}, object detection \cite{gene2020fruit}, and segmentation \cite{nasir2021analysis, pishgar2021redeca}. Common agricultural prediction tasks include weed, disease, plant, and fruit classification, localization, and/or segmentation \cite{lashgari2020fusion,lu2020survey}. In recent years, convolutional neural networks (CNNs) have become almost ubiquitous for these tasks due to their strong performance on image-based datasets \cite{kamilaris2018review}. 

\begin{figure}[ht!]
\begin{center}
\centerline{\includegraphics[width=1.0\linewidth]{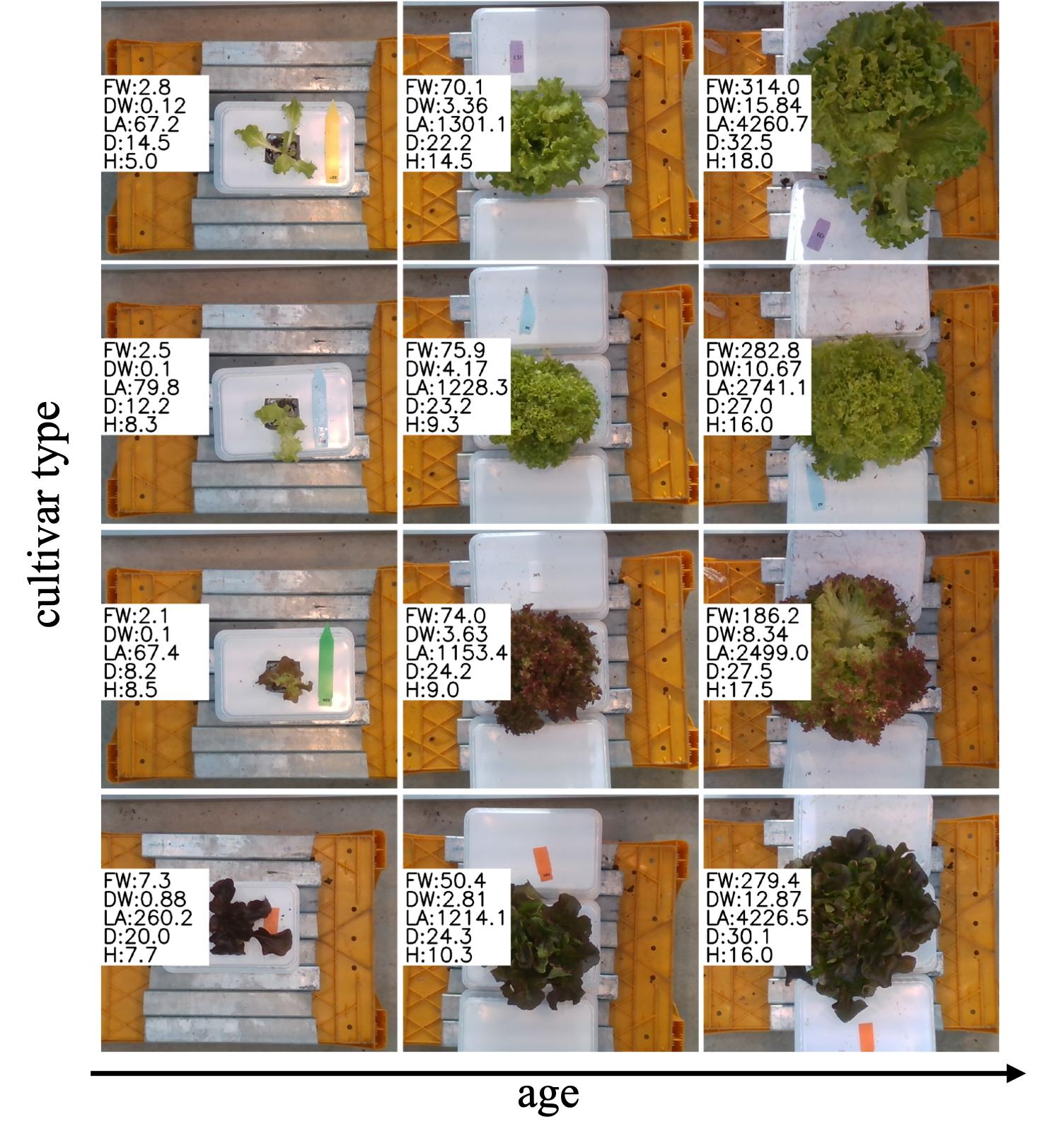}}
\caption{Data examples from the 2021 Autonomous Greenhouse Challenge Computer Vision dataset which contained four cultivars of lettuce imaged at various growth stages \cite{autonomousgreenhousedataset}.}
\label{fig:rawdata}
\end{center}
\vspace{-3.5em}
\end{figure}

\begin{figure*}[htp]
\begin{center}
\includegraphics[width=1.0\linewidth]{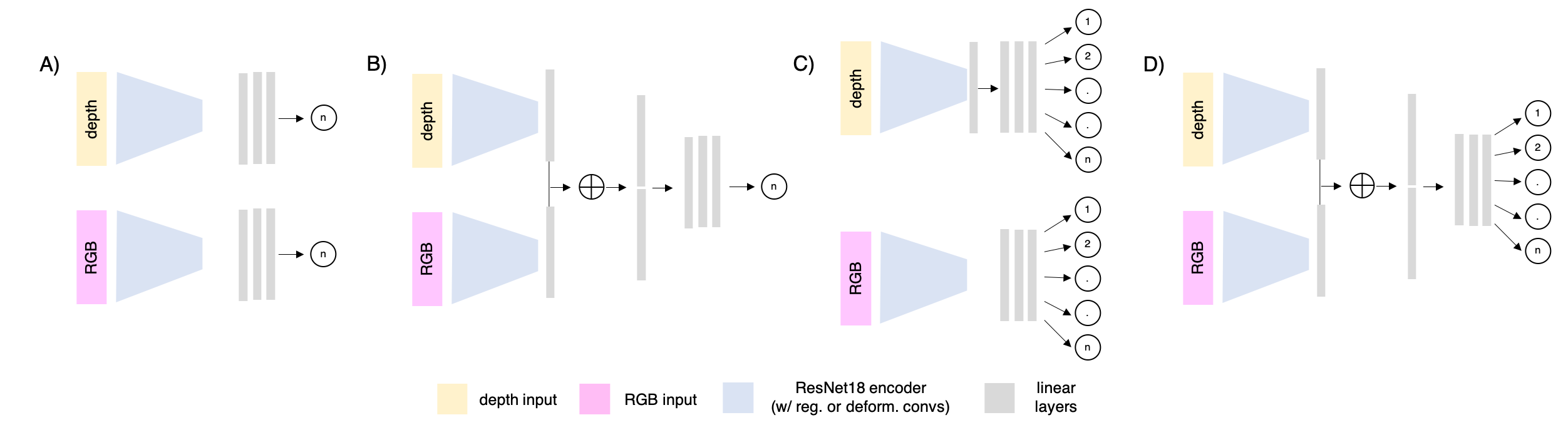}
\caption{Model architectures consisting of (A) single-input, single-output (SISO), (B) multi-input, single-output (MISO), (C) single-input, multi-output (SIMO), and (D) multi-input, multi-ouput (MIMO).}
\label{fig:architecture}
\end{center}
\vspace{-1em}
\end{figure*}

Regression is another highly relevant task for breeding and agriculture. CNNs can be effective at regressing plant traits from aerial and ground-based RGB imagery \cite{nevavuori2020crop, kestur2019mangonet, van2020crop, mortensen2019oil}. While some studies exist, CNN-based regression is far less common; possibly because CNN-based regression requires some architectural and loss modification, and/or continuous trait value measurements as ground-truth labels. Even less common are examples using CNNs to simultaneously predict multiple trait outputs from single or multiple sensor inputs \cite{SANDHU_WHEAT}. Although multi-input, multi-output trait regression is a common problem in agriculture \cite{kestur2019mangonet, van2020crop}, there is a lack of easily accessible code and frameworks for this purpose \cite{ kestur2019mangonet}.

Although CNNs have performed well on vision tasks, including regression, they suffer from some major drawbacks. First, convolution layers struggle to model complicated geometric transformations without exhaustive image augmentation such as flipping and rotating \cite{dai2017deformable}, scale segregation \cite{hu2017finding}, or introducing more architectural changes \cite{lin2017feature} which turn increases convergence time, training iterations and model complexity. Moreover, resize augmentations, which are useful to help with scale invariance, are not straightforward with image-wise plant trait labels since their effect on variables such as fresh weight can be hard to quantify. Another major drawback of the fixed-size kernels in conventional CNNs is their inability to modify the size and scale of their receptive field as they rely on pooling layers with fixed ratios \cite{dai2017deformable}. Deformable Convolutions solve this problem by independently generating offsets using a separate convolution layer prior to the main convolution operation. These offsets are then used to modify the location of the pixels which are input to the main convolution kernel, meaning the pixels in the output are no longer required to be dependent on neighboring pixels in the input. The geometric variability in the size and shape of plants and fruit and the high correlation of these geometric features with continuous trait outputs (e.g. fresh weight, leaf area, and dry weight) makes an adaptive receptive field, greater scale invariance, and abstraction of pixel and feature locations desirable in a model. Accordingly, we hypothesized that deformable convolution layers can be used to replace standard convolution layers, potentially improving prediction for tasks with geometric relevance.

Here, we present a relatively simple CNN model that accepts multiple sensor inputs and predicts multiple continuous trait outputs – i.e. a multi-input, multi-output CNN (MIMO-CNN). Furthermore, we introduce deformable convolutional layers into this network architecture (MIMO-DCNN) to enable the model to adaptively adjust its receptive field, model complicated variable geometric transformations in the data, and fine-tune the continuous trait outputs. We demonstrate state-of-the-art performance on the 2021 Autonomous Greenhouse Image Processing Challenge, and also provide a relatively simple pipeline that can be easily modified to accept custom inputs and outputs for other agricultural and plant breeding datasets.

\section{Methods}

All data used in this study was obtained from the Autonomous Greenhouse Challenge Dataset \cite{autonomousgreenhousedataset}. First, images were cropped manually to a size small enough to remove extraneous information but also left big enough to avoid cutting off large crops towards the end of the growing stages as shown in Figure \ref{fig:rawdata}. We defined this cropped region as 200 to 900 in y and 650 to 1450 in x, creating a 700 x 800 pixel image. Aligned with standard practice, and to set the RGB and depth values within a similar range, the images were normalized (i.e. centered and scaled) channel-wise. The means and standard deviation of each channel were calculated for the cropped train set alone and used for normalization in both training and inference. Random horizontal and vertical flip, random rotation, and random shift was used as training augmentations. 

Two sampling methods were tried during the training process. Each image was assigned to a bin based on fresh weight magnitude and sampled inversely proportional to the number of images in that bin (i.e. frequency). Stratified sampling based on plant variety was also tried. However, neither of the two methods yielded superior performance in comparison to a simple random sampling method without replacement. As defined by the 2021 Autonomous Greenhouse organizers, the test set of 50 images was separated and held out during the training process. The remaining 338 images were randomly split with a train to validation ratio of 0.75 to 0.25. This resulted in a validation set of 68 images and a train set of 270 images. Normalized Mean Squared Error (NMSE) was used a the loss for training,

\begin{equation} \label{eq:1}
\begin{gathered}
\mathit{NMSE}=\sum_{j=1}^m \frac{\sum_{i=1}^n (gt_{ij}-p_{ij})^2}{\sum_{i=1}^n (gt_{ij})^2} \\[1ex]
\end{gathered}
\end{equation}

where gt is the ground truth value, p is the predicted value, n is the number of images, i is the batch dimension, m is the number of traits/outputs and j is the output dimension. NMSE was chosen because it normalizes the MSE of the prediction and ground truth by dividing it by squared ground-truth values for each output. This allows a single model to be trained to predict multiple distinct outputs with different units without the need for output normalization. This also allows for error within a single variable to be penalized according to its proportion to the actual ground truth value. 

As shown in Figure \ref{fig:architecture}, we tried numerous model architectures that varied 1) single (SI-) versus multiple inputs (MI-), 2) single (-SO) versus multi-output (-MO), and 3) standard (-CNN) versus deformable convolutions (-DCNN). This resulted in a total of 18 model types for each standard and deformable convolution approaches, since (a) the SISO approach consisted of 10 total sub-models each trained on a single-input and single-output, (b) the SIMO approach consisted of two sub-models that accepted each input (RGB or Depth), (c) the MISO models consisted of five total sub-models each trained on a single output and all inputs, and (d) the MIMO approach consisted of one model, trained on all inputs and all outputs. For the multi-input models, we tried combining the RGB and depth inputs in a single model using early and mid-fusion approaches. Early-fusion was done by changing the initial convolution layer to accept 4 channel inputs rather than 3, but yielded worse results and was not examined further. Our mid-fusion approach consisted of two pretrained ResNet18 \cite{he2016deep} encoders, one for depth and the other for RGB. An additional convolutional layer to go from 1 to 3 channels was added to the beginning of the depth encoder to act as a color mapping layer. The outputs of each encoder are concatenated before applying non-linear transformations and linear operations to get to the output. A second architectural modification was implemented in which all standard convolution layers were replaced with deformable convolutional layers \cite{dai2017deformable}. The pre-trained weights of each original convolution layer were copied into the new deformable convolution layer. The number of offsets was set to 3 for the input layer and 8 for all other deformable convolution layers.

Adam optimizer was used with a learning rate of 0.0005 and no learning rate scheduling. All models were trained using the NMSE Loss function shown in equation 1 for consistency. The model was evaluated using the NMSE on the final test set for all traits, along with mean squared error (MSE) for each individual trait (see \ref{table:1}). All models were trained using the PyTorch library \cite{NEURIPS2019_9015} in Python.

\begin{table*}[t]
\begin{center}
\begin{tabular}{|c|c|c|c|c|c|c|c|c|c|c|c|c|}

\hline
\multicolumn{13}{|c|}{\textbf{Standard Convolutions vs Deformable Convolutions}} \\
\hline
\multicolumn{1}{|c|}{} &
\multicolumn{10}{c|}{MSE} &
\multicolumn{2}{c|}{NMSE}\\
\hline
& \multicolumn{2}{c|}{Fresh Wt} & \multicolumn{2}{c|}{Dry Wt} & \multicolumn{2}{c|}{Height} & \multicolumn{2}{c|}{Diameter} & \multicolumn{2}{c|}{Leaf Area} & \multicolumn{2}{c|}{All} \\

\hline
 & CNN & DCNN & CNN & DCNN & CNN & DCNN & CNN & DCNN & CNN & DCNN & CNN & DCNN \\
\hline
MIMO &8.97e2& \textbf{4.22e2}& 1.21& 0.97& 3.05& 3.07& 5.31& 6.63& 1.19e5& 7.52e4& 0.092& \textbf{0.068} \\
\hline
MISO &9.41e2& 6.52e2& 1.22& \textbf{0.78}& \textbf{2.02}& 2.69& \textbf{4.96}& 5.69 & 1.25e5 & \textbf{7.43e4}& 0.088& 0.069\\
\hline
SIMO-R &1.00e3& 9.82e2& 1.44& 1.09 & 3.31& 3.36& 4.99& 6.30& 1.16e5& 1.54e5& 0.099& 0.102 \\
\hline
SIMO-D &1.36e3& 8.46e2 & 1.55 & 1.53& 3.43 &4.64& 8.20 & 6.59& 9.45e4 & 1.07e5& 0.114& 0.104 \\
\hline
SISO-R & 1.05e3& 1.07e3 & 1.26& 0.89 & 3.01 & 2.71& 5.74 &  5.52& 1.19e5 & 1.35e5& 0.098& 0.093 \\
\hline
SISO-D & 8.98e2& 8.72e2& 1.68& 1.58& 3.00& 2.34& 6.89& 7.61& 9.72e4& 1.01e5& 0.098& 0.094 \\
\hline

\end{tabular}
\end{center}
\caption{Model performance of ResNet18-based regression models using standard and deformable convolutions with multi-input multi-output (MIMO), multi-input single-output (MISO), single-input multi-output RGB (SIMO-R), single-input multi-output depth (SIMO-D), single-input single-output RGB (SISO-R), single-input single-output Depth (SISO-D). For reference the leaderboard score was 0.081 for the 2021 Autonomous Greenhouse Challenge Computer Vision competition.}
\label{table:1}
\vspace{-1em}
\end{table*}

\begin{figure}[ht]
\begin{center}
\includegraphics[width=1.0\linewidth]{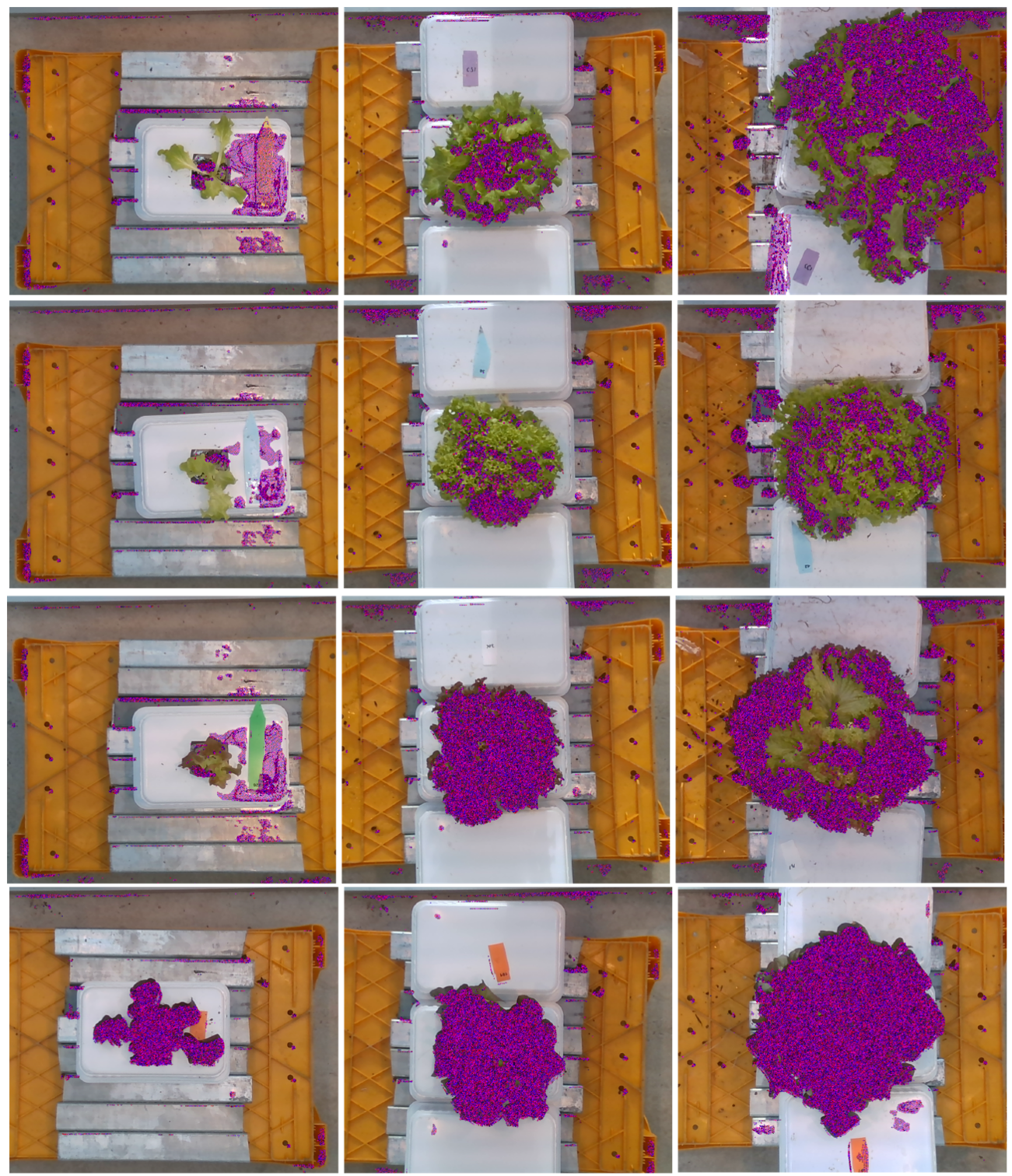}
\caption{Locations of the strong offsets produced from the first layer. Strong offsets are defined as offsets whose magnitude is greater than 3 and are indicated by the red, blue and violet points}
\label{fig:offsetdata}
\end{center}
\vspace{-1.5em}
\end{figure}

After model training, the output of the offset convolution operation in the earliest deformable convolution operation in the MIMO model was analyzed. We analyzed the earlier layers to avoid the "spatial abstraction" problem discussed previously. Although various approaches were attempted, the simplest method to visualize the offsets was to filter out "weaker offsets". We defined weaker offsets as having magnitude less then 3 px, from the original sampling location, and only showed offsets that were 3 px or more away from their standard sampling location. The red, blue, and violent point in Figure\ref{fig:offsetdata} show the location of these "strong" offsets for 4 kernel points. The simple abundance of points and artifacts made the offsets hard to visualize when too many offsets for too many kernel points were plotted at once.

\section{Results and Discussion}

We found that the MIMO-DCNN and MISO-DCNN approaches achieved the best performance in terms of NMSE on the 2021 Autonomous Greenhouse Image Processing Online Challenge Dataset, in addition to converging much earlier than all other models tested. Although the overall NMSE of both approaches were similar, the per variable error breakdown in Table \ref{table:1} shows that the MIMO-DCNN approach performed much better on fresh weight while the MISO-DCNN approach slightly outperformed MIMO-DCNN in all other variables. SIMO approaches performed worse across the board with SIMO-CNN Depth performing the worst.

The MIMO-DCNN and MISO-DCNN models outperformed the other approaches, with the MIMO-DCNN model outperforming the MISO-DCNN models, showing that multiple inputs are crucial in estimating these traits with high accuracy. However, certain variables could be over fitting when additional non-essential inputs are given, making ablation studies useful. For example, the results for the standard CNN models show that the single-input depth models were consistently better at predicting leaf area than multi-input models. While this trend did not carry over when deformable convolutions were tested, this suggests that not all inputs are likely essential for optimal performance when predicting certain variables.

As shown in Table \ref{table:1}, single- and multi-output models performed very similarly when multiple inputs were available. However, most of the variables in the MISO-DCNN approach were predicted with the best performance with exception of fresh weight, which performed better in the MIMO-DCNN approach. This may suggest that the mutual feature sharing of the multiple outputs may be beneficial for some variables but detrimental for others. In this case, fresh weight may have represented a comprehensive trait which was correlated with the geometric traits (i.e. larger plants likely had a higher fresh weight), such that training multi-output models on all traits simultaneously resulted in improved performance. In both the deformable and standard convolution models, single-output models seemed to a have higher advantage when inputs were ablated, suggesting that mutual feature sharing can be detrimental to many variables when only RGB or depth information is present. However, among the single output models, the fresh weight trait was still best predicted with a multi-input approach. Nevertheless, the abundance of single-input, RGB only datasets identified by \cite{lu2020survey} could make single-output not only useful but essential in estimating multiple traits with optimal performance.

Overall, models with deformable convolutions outperformed those with standard convolutions, especially when all inputs were given to the model. In comparison to the standard convolution models, the deformable convolution models seemed to better utilize information from both inputs, since the increase in performance between single- and multi-input models was substantially higher in the DCNN. Deformable convolutions performed much better in predicting fresh weight, dry weight and leaf area, showing that these traits benefit from having a adaptive receptive field and spatial abstraction of pixels and feature locations. Figure 3 shows the locations of the strongest offset for 4 kernel points in the earliest layers. We observe these kernel points specializing in extracting features in a more geometrically invariant way, since much of the offset points fall within the edges the plant telling us the sampling location of these kernel points mostly lie within the plant. The pixel values inside (and outside) the plants are then transformed into a more abstract coordinate plane since the receptive fields are not constant, which is likely what allows deformable convolutions to perform better. Importantly, this method fails to visually depict saliency, since there can be many artifacts outside of the features of interest \cite{Zhang2021RevisitingTD} and the gradients are not back-propagated to weight the layer's activations as in Grad-CAM \cite{8237336} and other similar approaches. Further, multi-input CNN methods still outperformed the DCNN methods for height and diameter. This could due to the fact that height and diameter information are relatively simpler to estimate and do not require complicated geometric transformations and spatial abstractions to model. If true, a simple set of kernels to determine the furthest edges in the height/width or depth dimensions can be used to determine these traits, causing the increased parameterization of the deformable convolution module \cite{dai2017deformable} to overfit. Simply put, the loss/abstraction of feature locations could be detrimental in cases where basic geometric measurement such as height or diameter need to be estimated.  Therefore, whether the outputs that are being estimated would benefit from higher spatial abstraction and receptive field adaptability when considering the use of deformable convolutions is an open question.

\section{Conclusion}
In this work we provide a simple, yet effective CNN regression approach to directly predict multiple plant traits from multiple sensor inputs. We test the performance of this method using the 2021 Autonomous Greenhouse Challenge Computer Vision dataset, demonstrating a 16\% reduction in NMSE compared to the top leaderboard score. Additionally, we conducted ablation tests to examine the importance of using multi-input RGB and depth information and multi-output trait data with a single or multiple models. Future work could test how more flexible methods, such as deformable convolutions V2 \cite{zhu2019deformable} and other spatial attention mechanisms \cite{Zhu_2019_ICCV}, affect performance on multi-input, multi-output regression problems. The technique and codebase developed here should be broadly useful for plant breeders and agricultural producers interested in predicting multiple traits from multiple sensors. 

\section{Acknowledgments}
This project was partly supported by the USDA AI Institute for Next Generation Food Systems (AIFS), USDA award number 2020-67021-32855.

{\small
\bibliography{Main}

\begin{thebibliography}{28}
\providecommand{\natexlab}[1]{#1}

\bibitem[{Akbar et~al.(2016)Akbar, Chattopadhyay, Elfiky, and
  Kak}]{akbar2016novel}
Akbar, S.~A.; Chattopadhyay, S.; Elfiky, N.~M.; and Kak, A. 2016.
\newblock A novel benchmark RGBD dataset for dormant apple trees and its
  application to automatic pruning.
\newblock In \emph{Proceedings of the IEEE Conference on Computer Vision and
  Pattern Recognition Workshops}, 81--88.

\bibitem[{Dai et~al.(2017)Dai, Qi, Xiong, Li, Zhang, Hu, and
  Wei}]{dai2017deformable}
Dai, J.; Qi, H.; Xiong, Y.; Li, Y.; Zhang, G.; Hu, H.; and Wei, Y. 2017.
\newblock Deformable convolutional networks.
\newblock In \emph{Proceedings of the IEEE international conference on computer
  vision}, 764--773.

\bibitem[{Elsherbiny et~al.(2021)Elsherbiny, Zhou, Feng, and
  Qiu}]{elsherbiny2021integration}
Elsherbiny, O.; Zhou, L.; Feng, L.; and Qiu, Z. 2021.
\newblock Integration of Visible and Thermal Imagery with an Artificial Neural
  Network Approach for Robust Forecasting of Canopy Water Content in Rice.
\newblock \emph{Remote Sensing}, 13(9): 1785.

\bibitem[{Gen{\'e}-Mola et~al.(2020)Gen{\'e}-Mola, Sanz-Cortiella, Rosell-Polo,
  Morros, Ruiz-Hidalgo, Vilaplana, and Gregorio}]{gene2020fruit}
Gen{\'e}-Mola, J.; Sanz-Cortiella, R.; Rosell-Polo, J.~R.; Morros, J.-R.;
  Ruiz-Hidalgo, J.; Vilaplana, V.; and Gregorio, E. 2020.
\newblock Fruit detection and 3D location using instance segmentation neural
  networks and structure-from-motion photogrammetry.
\newblock \emph{Computers and Electronics in Agriculture}, 169: 105165.

\bibitem[{Gené-Mola et~al.(2019)Gené-Mola, Vilaplana, Rosell-Polo, Morros,
  Ruiz-Hidalgo, and Gregorio}]{GENEMOLA2019104289}
Gené-Mola, J.; Vilaplana, V.; Rosell-Polo, J.~R.; Morros, J.-R.; Ruiz-Hidalgo,
  J.; and Gregorio, E. 2019.
\newblock KFuji RGB-DS database: Fuji apple multi-modal images for fruit
  detection with color, depth and range-corrected IR data.
\newblock \emph{Data in Brief}, 25: 104289.

\bibitem[{Haug and Ostermann(2014)}]{haug2014crop}
Haug, S.; and Ostermann, J. 2014.
\newblock A crop/weed field image dataset for the evaluation of computer vision
  based precision agriculture tasks.
\newblock In \emph{European conference on computer vision}, 105--116. Springer.

\bibitem[{He et~al.(2016)He, Zhang, Ren, and Sun}]{he2016deep}
He, K.; Zhang, X.; Ren, S.; and Sun, J. 2016.
\newblock Deep residual learning for image recognition.
\newblock In \emph{Proceedings of the IEEE conference on computer vision and
  pattern recognition}, 770--778.

\bibitem[{Hemming et~al.(2021)Hemming, de~Zwart, Elings, bijlaard monique,
  Marrewijk, and Petropoulou}]{autonomousgreenhousedataset}
Hemming, S.~S.; de~Zwart, H.~F.; Elings, A.~A.; bijlaard monique; Marrewijk,
  B., van; and Petropoulou, A. 2021.
\newblock 3rd Autonomous Greenhouse Challenge: Online Challenge Lettuce Images.
\newblock 4TU.ResearchData, \url{https://doi.org/10.4121/15023088.v1}.

\bibitem[{Hu and Ramanan(2017)}]{hu2017finding}
Hu, P.; and Ramanan, D. 2017.
\newblock Finding tiny faces.
\newblock In \emph{Proceedings of the IEEE conference on computer vision and
  pattern recognition}, 951--959.

\bibitem[{Kamilaris and Prenafeta-Bold{\'u}(2018)}]{kamilaris2018review}
Kamilaris, A.; and Prenafeta-Bold{\'u}, F.~X. 2018.
\newblock A review of the use of convolutional neural networks in agriculture.
\newblock \emph{The Journal of Agricultural Science}, 156(3): 312--322.

\bibitem[{Kestur, Meduri, and Narasipura(2019)}]{kestur2019mangonet}
Kestur, R.; Meduri, A.; and Narasipura, O. 2019.
\newblock MangoNet: A deep semantic segmentation architecture for a method to
  detect and count mangoes in an open orchard.
\newblock \emph{Engineering Applications of Artificial Intelligence}, 77:
  59--69.

\bibitem[{Kusumam et~al.(2016)Kusumam, Krajn{\'\i}k, Pearson, Cielniak, and
  Duckett}]{kusumam2016can}
Kusumam, K.; Krajn{\'\i}k, T.; Pearson, S.; Cielniak, G.; and Duckett, T. 2016.
\newblock Can you pick a broccoli? 3D-vision based detection and localisation
  of broccoli heads in the field.
\newblock In \emph{2016 IEEE/RSJ International Conference on Intelligent Robots
  and Systems (IROS)}, 646--651. IEEE.

\bibitem[{Lashgari, Imanmehr, and Tavakoli(2020)}]{lashgari2020fusion}
Lashgari, M.; Imanmehr, A.; and Tavakoli, H. 2020.
\newblock Fusion of acoustic sensing and deep learning techniques for apple
  mealiness detection.
\newblock \emph{Journal of Food Science and Technology}, 57(6): 2233--2240.

\bibitem[{Lin et~al.(2017)Lin, Doll{\'a}r, Girshick, He, Hariharan, and
  Belongie}]{lin2017feature}
Lin, T.-Y.; Doll{\'a}r, P.; Girshick, R.; He, K.; Hariharan, B.; and Belongie,
  S. 2017.
\newblock Feature pyramid networks for object detection.
\newblock In \emph{Proceedings of the IEEE conference on computer vision and
  pattern recognition}, 2117--2125.

\bibitem[{Lottes et~al.(2018)Lottes, Behley, Milioto, and
  Stachniss}]{lottes2018fully}
Lottes, P.; Behley, J.; Milioto, A.; and Stachniss, C. 2018.
\newblock Fully convolutional networks with sequential information for robust
  crop and weed detection in precision farming.
\newblock \emph{IEEE Robotics and Automation Letters}, 3(4): 2870--2877.

\bibitem[{Lu and Young(2020)}]{lu2020survey}
Lu, Y.; and Young, S. 2020.
\newblock A survey of public datasets for computer vision tasks in precision
  agriculture.
\newblock \emph{Computers and Electronics in Agriculture}, 178: 105760.

\bibitem[{Mortensen et~al.(2019)Mortensen, Skovsen, Karstoft, and
  Gislum}]{mortensen2019oil}
Mortensen, A.~K.; Skovsen, S.; Karstoft, H.; and Gislum, R. 2019.
\newblock The Oil Radish Growth Dataset for Semantic Segmentation and Yield
  Estimation.
\newblock In \emph{CVPR Workshops}, 2703--2710.

\bibitem[{Nasir, Azman, and Arsat(2021)}]{nasir2021analysis}
Nasir, N. U.~M.; Azman, M.~I.; and Arsat, Z.~A. 2021.
\newblock Analysis for the Estimation of Harumanis Mango Ripeness Guide.
\newblock \emph{Advances in Agricultural and Food Research Journal}, 2(1).

\bibitem[{Nevavuori et~al.(2020)Nevavuori, Narra, Linna, and
  Lipping}]{nevavuori2020crop}
Nevavuori, P.; Narra, N.; Linna, P.; and Lipping, T. 2020.
\newblock Crop yield prediction using multitemporal uav data and
  spatio-temporal deep learning models.
\newblock \emph{Remote Sensing}, 12(23): 4000.

\bibitem[{Paszke et~al.(2019)Paszke, Gross, Massa, Lerer, Bradbury, Chanan,
  Killeen, Lin, Gimelshein, Antiga, Desmaison, Kopf, Yang, DeVito, Raison,
  Tejani, Chilamkurthy, Steiner, Fang, Bai, and Chintala}]{NEURIPS2019_9015}
Paszke, A.; Gross, S.; Massa, F.; Lerer, A.; Bradbury, J.; Chanan, G.; Killeen,
  T.; Lin, Z.; Gimelshein, N.; Antiga, L.; Desmaison, A.; Kopf, A.; Yang, E.;
  DeVito, Z.; Raison, M.; Tejani, A.; Chilamkurthy, S.; Steiner, B.; Fang, L.;
  Bai, J.; and Chintala, S. 2019.
\newblock PyTorch: An Imperative Style, High-Performance Deep Learning Library.
\newblock In Wallach, H.; Larochelle, H.; Beygelzimer, A.; d\textquotesingle
  Alch\'{e}-Buc, F.; Fox, E.; and Garnett, R., eds., \emph{Advances in Neural
  Information Processing Systems 32}, 8024--8035. Curran Associates, Inc.

\bibitem[{Pishgar et~al.(2021)Pishgar, Issa, Sietsema, Pratap, and
  Darabi}]{pishgar2021redeca}
Pishgar, M.; Issa, S.~F.; Sietsema, M.; Pratap, P.; and Darabi, H. 2021.
\newblock REDECA: A Novel Framework to Review Artificial Intelligence and Its
  Applications in Occupational Safety and Health.
\newblock \emph{International Journal of Environmental Research and Public
  Health}, 18(13): 6705.

\bibitem[{Sa et~al.(2017)Sa, Chen, Popovi{\'c}, Khanna, Liebisch, Nieto, and
  Siegwart}]{sa2017weednet}
Sa, I.; Chen, Z.; Popovi{\'c}, M.; Khanna, R.; Liebisch, F.; Nieto, J.; and
  Siegwart, R. 2017.
\newblock weednet: Dense semantic weed classification using multispectral
  images and mav for smart farming.
\newblock \emph{IEEE Robotics and Automation Letters}, 3(1): 588--595.

\bibitem[{Sandhu et~al.(2021)Sandhu, Lozada, Zhang, Pumphrey, and
  Carter}]{SANDHU_WHEAT}
Sandhu, K.~S.; Lozada, D.~N.; Zhang, Z.; Pumphrey, M.~O.; and Carter, A.~H.
  2021.
\newblock Deep Learning for Predicting Complex Traits in Spring Wheat Breeding
  Program.
\newblock \emph{Frontiers in Plant Science}, 11: 2084.

\bibitem[{Selvaraju et~al.(2017)Selvaraju, Cogswell, Das, Vedantam, Parikh, and
  Batra}]{8237336}
Selvaraju, R.~R.; Cogswell, M.; Das, A.; Vedantam, R.; Parikh, D.; and Batra,
  D. 2017.
\newblock Grad-CAM: Visual Explanations from Deep Networks via Gradient-Based
  Localization.
\newblock In \emph{2017 IEEE International Conference on Computer Vision
  (ICCV)}, 618--626.

\bibitem[{Van~Klompenburg, Kassahun, and Catal(2020)}]{van2020crop}
Van~Klompenburg, T.; Kassahun, A.; and Catal, C. 2020.
\newblock Crop yield prediction using machine learning: A systematic literature
  review.
\newblock \emph{Computers and Electronics in Agriculture}, 177: 105709.

\bibitem[{Zhang et~al.(2021)Zhang, Xie, Luo, and Cao}]{Zhang2021RevisitingTD}
Zhang, Y.; Xie, Y.; Luo, L.; and Cao, F. 2021.
\newblock Revisiting the Deformable Convolution by Visualization.
\newblock In \emph{ICPRAM}.

\bibitem[{Zhu et~al.(2019{\natexlab{a}})Zhu, Cheng, Zhang, Lin, and
  Dai}]{Zhu_2019_ICCV}
Zhu, X.; Cheng, D.; Zhang, Z.; Lin, S.; and Dai, J. 2019{\natexlab{a}}.
\newblock An Empirical Study of Spatial Attention Mechanisms in Deep Networks.
\newblock In \emph{Proceedings of the IEEE/CVF International Conference on
  Computer Vision (ICCV)}.

\bibitem[{Zhu et~al.(2019{\natexlab{b}})Zhu, Hu, Lin, and
  Dai}]{zhu2019deformable}
Zhu, X.; Hu, H.; Lin, S.; and Dai, J. 2019{\natexlab{b}}.
\newblock Deformable convnets v2: More deformable, better results.
\newblock In \emph{Proceedings of the IEEE/CVF Conference on Computer Vision
  and Pattern Recognition}, 9308--9316.

\end{thebibliography}
}

\end{document}